\documentclass[sigconf,nonacm]{acmart}
\usepackage{caption}
\usepackage{multirow}
\usepackage{xcolor}
\usepackage{graphicx}
\usepackage{epstopdf}
\usepackage{url}
\usepackage{comment}
\usepackage{graphicx}
\usepackage{caption}
\usepackage{subcaption}
\AtBeginDocument{%
  \providecommand\BibTeX{{%
  \normalfont 
  \kern-0.1em{\scshape i\kern-0.1em b}\kern-0.1em\TeX}
  }}
\begin{document}

\title{Enhanced Local Explainability and Trust Scores with Random Forest Proximities}


% \author{Anonymous1}
% % \email{Anonymous@}
% \affiliation{%
% 	% \institution{Anonymous}
% 	% \city{New York}
% 	% \state{NY}
% 	\country{Anonymous}
% }
% \author{Anonymous2}
% % \email{Anonymous@}
% \affiliation{%
% 	% \institution{Anonymous}
% 	% \city{New York}
% 	% \state{NY}
% 	\country{Anonymous}
% }
% \author{Anonymous3}
% % \email{Anonymous@}
% \affiliation{%
% 	% \institution{Anonymous}
% 	% \city{New York}
% 	% \state{NY}
% 	\country{Anonymous}
% }
% \author{Anonymous4}
% % \email{Anonymous@}
% \affiliation{%
% 	% \institution{Anonymous}
% 	% \city{New York}
% 	% \state{NY}
% 	\country{Anonymous}
% }
% \author{Anonymous5}
% % \email{Anonymous@}
% \affiliation{%
% 	% \institution{Anonymous}
% 	% \city{New York}
% 	% \state{NY}
% 	\country{Anonymous}
% }
%\author{Anonymous Authors}
%\affiliation{%
%	\institution{Anonymous}
%	% \city{New York}
%	% \state{NY}
%	\country{Anonymous}
%}
% \renewcommand{\shortauthors}{Anonymous et al.}

% \begin{comment}
    
\author{Joshua Rosaler}
\email{joshua.rosaler@blackrock.com}
\affiliation{%
\institution{BlackRock, Inc.}
\city{New York}
\state{NY}
\country{USA}
}
 \author{Dhruv Desai}
 \email{dhruv.desai@blackrock.com}
 \affiliation{%
   \institution{BlackRock, Inc.}
  \city{New York}
   \state{NY}
   \country{USA}
 }
 \author{Bhaskarjit Sarmah}
 \email{bhaskarjit.sarmah@blackrock.com}
 \affiliation{%
   \institution{BlackRock, Inc.}
   \city{Gurgaon}
   \country{India}
 }
 \author{Dimitrios Vamvourellis}
 \email{dimitrios.vamvourellis@blackrock.com}
 \affiliation{%
   \institution{BlackRock, Inc.}
   \city{New York}
   \state{NY}
   \country{USA}
 }

 \author{Deran Onay}
 \email{deran.onay@blackrock.com}
 \affiliation{%
   \institution{BlackRock, Inc.}
   \city{New York, NY}
   \country{USA}
 }

 \author{Stefano Pasquali}
 \email{stefano.pasquali@blackrock.com}
 \affiliation{%
   \institution{BlackRock, Inc.}
   \city{New York, NY}
   \country{USA}
   }
 \author{Dhagash Mehta}
 \email{dhagash.mehta@blackrock.com}
 \affiliation{%
   \institution{BlackRock, Inc.}
   \city{New York, NY}
   \country{USA}
 }

 \renewcommand{\shortauthors}{Rosaler et al.}
% \end{comment}

\begin{abstract}
We initiate a novel approach to explain the predictions and out of sample performance of random forest (RF) regression and classification models by exploiting the fact that any RF can be mathematically formulated as an adaptive weighted K nearest-neighbors model. Specifically, we employ a recent result that, for both regression and classification tasks, any RF prediction can be rewritten exactly as a weighted sum of the training targets, where the weights are RF proximities between the corresponding pairs of data points. We show that this linearity facilitates a local notion of explainability of RF predictions that generates attributions for any model prediction across observations in the training set, and thereby complements established feature-based methods like SHAP, which generate attributions for a model prediction across input features. We show how this proximity-based approach to explainability can be used in conjunction with SHAP to explain not just the model predictions, but also out-of-sample performance, in the sense that proximities furnish a novel means of assessing when a given model prediction is more or less likely to be correct. We demonstrate this approach in the modeling of US corporate bond prices and returns in both regression and classification cases. 
\end{abstract}

\begin{comment}
\begin{CCSXML}
<ccs2012>
 <concept>
  <concept_id>10010520.10010553.10010562</concept_id>
  <concept_desc>Computer systems organization~Embedded systems</concept_desc>
  <concept_significance>500</concept_significance>
 </concept>
 <concept>
  <concept_id>10010520.10010575.10010755</concept_id>
  <concept_desc>Computer systems organization~Redundancy</concept_desc>
  <concept_significance>300</concept_significance>
 </concept>
 <concept>
  <concept_id>10010520.10010553.10010554</concept_id>
  <concept_desc>Computer systems organization~Robotics</concept_desc>
  <concept_significance>100</concept_significance>
 </concept>
 <concept>
  <concept_id>10003033.10003083.10003095</concept_id>
  <concept_desc>Networks~Network reliability</concept_desc>
  <concept_significance>100</concept_significance>
 </concept>
</ccs2012>
\end{CCSXML}

\ccsdesc[500]{Computer systems organization~Embedded systems}
\ccsdesc[300]{Computer systems organization~Redundancy}
\ccsdesc{Computer systems organization~Robotics}
\ccsdesc[100]{Networks~Network reliability}

\keywords{Company similarity, Natural Language Processing}

\end{comment}

\maketitle

\graphicspath{ {plots/plots_rf_explainability/} }

\section{Introduction}
With the growing dependence on machine learning (ML) in the financial domain, it is becoming increasingly necessary to gain transparency into the underlying mechanisms that drive the predictions of complex non-linear ``blackbox" machine learning models, including artificial neural networks \cite{goodfellow2016deep}, Gradient Boosted Machines (GBMs) \cite{friedman2001greedy}, Random Forests (RFs) \cite{breiman2001random}, and many others. However, there is no consensus definition of explainability for ML models that has been agreed upon by financial and ML domain experts and regulatory authorities, complicating the issue further \cite{doshi2017towards,molnar2020interpretable,burkart2021survey}.

Though global explainability methods such as permutation importance, partial dependence plots, Accumulated Local Effects plots, etc. have been extensively and traditionally used, recently some of the most popular paradigms for explainability for ML models have been focused on so-called %\textit{post hoc} 
\textit{local} explanability where the prediction of each of the data-points by the model is explained using a method such as SHapley Additive exPlanations (SHAP) \cite{lundberg2017unified,lundberg2020local}, Local Interpretable Model-Agnostic Explanations (LIME) \cite{ribeiro2016should}, DeepLift \cite{shrikumar2016not}, Anchors \cite{ribeiro2018anchors}, etc. 

Many of the local explainability methods provide ``feature-based" explanations in the sense that they rely on a linear attribution of the model's prediction for a given observation to different input features. The feature-based paradigm for explainability is most simply illustrated by linear regression, in that each term in a linear regression model constitutes a feature's additive contribution to the final prediction of the model. The most popular extension of this feature-based paradigm to non-linear models is the SHAP algorithm, which provides for a given observation a measure of the size and direction of each feature's additive contribution to the model's prediction for that observation \cite{lundberg2017unified}. Moreover, SHAP values are the unique feature attributions satisfying the natural criteria of full attribution, fairness, and monotonicity \cite{shapley1997value}. %LIME (Local Interpretable Model-Agnostic Explanations), on the other hand, provides an alternative method of feature-based attribution by fitting a linear regression model to simulated input-output pairs drawn from the model of interest, where the simulated inputs are constrained to lie in the vicinity of the test point. Other feature-based methods such as Anchor and Deeplift, rely on still other methods of assigning responsibility for the model predictions across dimensions of the input space. 

Less well known is the alternative ``instance-based" paradigm for explainability, which is based instead on a linear attribution of the model's prediction to different points in the model's training set. While the feature-based paradigm is most simply illustrated by linear regression, the instance-based paradigm is most simply illustrated by the K-nearest-neighbors (KNN) algorithm \cite{fix1989discriminatory}. Here, to explain why an out-of-sample prediction was off in a given case, one would look at the target labels of the K nearest neighbors and check that the target label for the test observation was anomalous relative to this subset of the training data that was used to generate the prediction. Moreover, if the nearest neighbors to the test point are mostly far away in feature space, or if the training error tends to be relatively larger in the test point's region of the feature space, this would help to explain cases where the KNN model is off.

%\textcolor{red}{@Josh, could you also please write a paragraph or two on how other local methods work, e.g., LIME, ANCHPR, Deeplift, etc.? We also need to put our method in wider perspective than just SHAP.}

%For tree-based algorithms, more specialized feature-based explainability methods such as TreeSHAP \cite{lundberg2020local} have been proposed. Along the same lines, 
In the present work, we exploit the fact that, by virtue of a certain mathematical equivalence, any ensemble tree-based model, and in particular, any random forest regressor or classifier, can be formulated exactly as an adaptive weighted KNN model - that is, as a weighted average of the training targets in which the weights vary over the feature space \cite{lin2006random} \cite{rhodes2023geometry}. We make use of this equivalence to provide an instance-based approach to explainability for the RF algorithm that complements feature-based approaches in the context of both RF regression and classification. Unlike in ordinary KNN, the notion of ``nearest" in the context of RF is now quantified in terms of a ``proximity" that can be extracted from the trained RF model, and from which a non-Euclidean distance metric on the RF model's feature space can be constructed. 
%Just as SHAP-based explanations focus attention on the subset of \textit{features} most important to generating the model prediction, the RF instance-based explanations focus attention on the subset of \textit{training observations} most relevant to generating the prediction. The proposed proximity-based local approach to explainability in the context of RF models therefore provides a useful complement to feature-based explanations like SHAP.

\section{Our Approach and Contributions}

First, the familiar SHAP-based method of attribution, as well as other feature-based methods of explanation, rest on a linear decomposition of the prediction $\hat{y}_{i}$:
\begin{equation}\label{eq:shap}
\hat{y}_{i} - \mathbb{E}(y) = s_{i,1}(x) + ... + s_{i,P}(x), 
\end{equation}
where $s_{i,j}(x)$ are local, input-dependent SHAP values for the data-point $i$ and feature $j$, and $j$ runs from $1,\dots, P$ with $P$ being the number of input features to the algorithm. $\mathbb{E}(y)$ is the unconditional expectation value of the target variable, which is the base prediction without any input features \cite{lundberg2017unified, lundberg2020local}.

On the other hand, our proposed method of model explanation starts with the interesting observation that any tree-based ensemble model, such as a random forest or gradient boosted machine, can be viewed as an adaptive weighted KNN model, as proved in Ref.\cite{lin2006random}. 

In the context of regression, this result entails that one can expand the prediction $\hat{y}_{i}$ of the model as a linear combination of target labels in the training dataset:

\begin{equation}\label{eq:RF_weighted_KNN}
\hat{y}_{i} = \textbf{k}_{i}(x) \cdot \textbf{y}_{train} = k_{i,1}(x) \ y_{train,1} \ + \ ... \ + \ k_{i,N}(x) \ y_{train,N},
\end{equation}
where $y_{train, j}$ is the ground truth target label for the $j^{th}$ training example, and $k_{i,j}(x)$ is the input-dependent weight or ``proximity" of the observation $j$ in the expansion for observation $i$ \cite{rhodes2023geometry}.

In the context of classification, it is not the predicted class, $c$, but rather the predicted probability $\rho_{i}^{c}$ of the random forest that is linearly expanded in terms of the training targets: 

\begin{equation}\label{eq:RF_weighted_KNN_clf}
\rho_{i}^{c} = k_{i,1}(x) \ I\left[y_{train,1}=c \right] + ... + k_{i,N}(x) \ I\left[y_{train,N}=c \right] 
\end{equation}

\noindent where $I[p]$ is the indicator function, equal to $1$ if $p$ is true and $0$ if $p$ is false; the predicted class is given by $\hat{y}_{i} = \text{argmax}_c \rho_{i}^{c}$. 
%\cite{rhodes2023geometry} show explicitly how the final class prediction of a random forest classifier can be expressed as the argmax of a linear expansion of the training labels of the form \label{eq:RF_weighted_KNN_clf}.   

The coefficients $k_{i,j}(x)$ correspond to one particular notion of proximity that can be defined for a tree-based ensemble. All such notions have in common that two points have higher proximity when they fall in the same leaf node for a higher proportion of decision trees in some tree-based ensemble (such as a random forest or gradient boosted machine). Recently, in Ref.~\cite{jeyapaulraj2022supervised}, a closely related notion of RF proximity was used to compute the similarity of corporate bonds for the purpose of identifying liquid substitutes in trading; in Ref.~\cite{desai2023quantifying}, RF proximities were used to identify outlier mutual funds within their assigned fund categories.

%\noindent  The specific form of the weights (i.e., proximities) $k_{i,j}$ in Eq.~(\ref{eq:RF_weighted_KNN}) depends on the family of tree-based models in question. 
\begin{comment}  
Breiman originally proposed the following definition of proximity for random forests \cite{breiman-cutler-blog}:
\begin{equation}\label{eq:breiman_prox}
Prox^{Breiman}(i,j) = \frac{1}{M} \sum_{T=1}^{M} I[j \in \mathcal{L}_{i}^{T}],
\end{equation}
where $M$ is the number of trees in the forest and $\mathcal{L}_{i}^{T}$ is the leaf of tree $T$ into which point $i$ falls.

A first, naive attempt at a general formula for the weights $k_{i,j}(x)$ in Eqs (\ref{eq:RF_weighted_KNN}) and (\ref{eq:RF_weighted_KNN_clf}) proposes expanding the RF prediction for point $i$ as the average over trees $T$ of the average over all training points $j$ that fall in the same leaf node as point $i$ of the targets $y_{train,j}$: 

\begin{equation}\label{eq:breiman_prox}
Prox^{Naive}(i,j) = \frac{1}{M} \sum_{T=1}^{M} \frac{1}{N_{i}^{T}}I[j \in \mathcal{L}_{i}^{T}],
\end{equation}
 where $M$ is the number of trees in the forest and $N_{i}^{T}$ is the number of training points in the leaf $\mathcal{L}_{i}^{T}$ of tree $T$ into which points $i$ and $j$ both fall. 
\end{comment}

Ref.~\cite{rhodes2023geometry} has shown that for random forests, the correct form of the expansion coefficents $k_{ij}$ appearing in Eqs (\ref{eq:RF_weighted_KNN}) and (\ref{eq:RF_weighted_KNN_clf}), which the authors
call Geometry and Accuracy Preserving (GAP) RF proximity, and which exactly recover the predictions of the random forest, is

\begin{equation} \label{GAP}
Prox^{GAP}(i,j) = \frac{1}{|S_{i}|} \sum_{t \in S_{i}} \frac{c_j(t)I[j \in J_{i}(t)]}{|M_{i}(t)|},
\end{equation}
where $S_{i}$ is the set of trees in the RF for which observation $i$ is out of bag, $M_i(t)$ is the multiset 
\footnote{Recall that a multiset is a generalization of the concept of a set, allowing for repetition among the elements of the set, where the number of repetitions of a unique element in the multiset is known as its multiplicity.}
of bagged points in the same leaf as $i$ in tree $t$, $J_{i}(t)$ is the corresponding set (i.e., without repetitions) of bagged points in the same leaf as $i$ in tree $t$, and $c_{j}(t)$ is the multiplicity of the index $j$ in the bootstrap sample. 
%Since this notion of proximity is asymmetric, it would need to be symmetrized before the corresponding similarity could be used to define a distance metric - e.g., through the symmetrization, i.e., $k^{GAP symm}_{i,j} = \frac{1}{2}(k^{GAP}_{i,j}+k^{GAP}_{j,i})$. 
%Here the expression for RF GAP proximity is shown to be correct for both regression and classification tasks, making it suitable for a complete analysis of explainability of RF predictions. 
Ref.~\cite{geertsema2023instancebased} have derived a method to compute the weights $k_{i,j}$ in (\ref{eq:RF_weighted_KNN}) for GBM regression; however, deriving the correct expressions for the weights for the GBM classifier remains an unsolved problem. %The authors of Ref.~\cite{geertsema2023instancebased}, however, passed over the case of RF claiming it as somewhat trivial by referring to the definition as in Eq.~(\ref{eq:breiman_prox}), and did not analyze the RF case in the paper. 

%Below, we explain our approach, implementation and analysis of the use of the GAP proximity for instance-based local explainability of RF predictions. To our knowledge, such a method of explainability for the RF algorithm has not been implemented or analysed before.

\subsection{The Proposed Methodology}\label{sec:methodology}
The purpose of our analysis is to show how GAP proximities can facilitate the explanation of RF model success or failure for a given observation based on its nearest neighbors in the training set, where ``nearest" is understood in the specific sense furnished by the proximities. 

The manner in which we employ the RF GAP proximities to explain success or failure of the model outputs is twofold. The first type of explanation, like SHAP or LIME, simply helps to explain why the model predicted what it did, irrespective of whether that prediction was close to or far from the ground truth label. The main difference in this case is that SHAP and LIME explanations attribute the prediction to different features while the GAP-based explanations attribute the prediction to different points in the training set. 

\subsection{Explaining Not Just Model Predictions, but Model Performance}

The second sense in which GAP proximities can help to explain the success or failure of the model goes further than mere attribution of the model predictions by actually explaining model performance against out-of-sample ground truth labels, by providing mathematical reasons why the model prediction was more or less likely to be far from the out-of-sample target label. Since each training point can also be associated with a certain level of performance of the model with respect to the train set labels, and since the proximities also provide a principled way to quantify how close a test point is to the bulk of the training data, proximities can explain why that model is more or less likely to perform well on a given test point based on the uncertainty in the target variable within this region of the feature space, and on a measure of how ``close" the test point is to bulk of training data.
%Assuming that the joint distribution of target and features does not shift substantially between the train and test sets, the training errors for the test point's nearest neighbors can be used to provide an estimate of the noise in the target variable in this region of the feature space. 

%\textcolor{red}{@Josh, its not clear what the proposed methodology is. Could you please write the actual proposed methodology for explanability for both regression and classification here? I think it may amount only to copy-paste certain paragraphs from the below sections.
%Also, probably the proposed methodology clearly written in step-by-step format or something may also help?}

\subsection{Applying SHAP and Proximity-Based Explanation Together}

For a given test point, we can use random forest proximities in conjunction with SHAP to explain both the model prediction and out-of-sample performance in a manner that is more comprehensive than either method on its own.  SHAP alone provides a way to explain the model predictions, but not \textit{performance} out of sample - that is, SHAP does not explain why the ground truth label was more or less likely to be close to the model prediction. 

When attempting to visualize the distribution within the input space of training points that are ``closest" to the test point and therefore contribute most to the prediction at that point, not all features are equally important. To visualize the proximity-weighted distribution of training points that contribute most to the prediction at the test point, we can use the local SHAP values at the test point to help us focus on just the one, two, or three most important features. For example, focusing on just the top two SHAP features, we can generate a scatterplot the proximity-weighted distribution of training points to visualize which training points contribute most to the model prediction (see Figs \ref{scatterplots_regression_badPred}, \ref{scatterplots_regression_goodPred}, \ref{scatterplots}, \ref{scatterplots_correct} ).

\section{Regression Use Case: Corporate Bond Price Discovery}

To illustrate the use of GAP proximities in the context of the RF regression, we focus on the task of price discovery in the US corporate bond market \cite{sommer2016liquidity,madhavan2022trading,fabozzi2011handbook}. The corporate bond market is substantially less liquid than the equity market, with many securities trading less than once per day, so there is often a lack of transparency as to the fair price of the security. By training an RF on trades across the universe of corporate bonds, we can make use of information from securities that trade more often to improve our predictions for the prices of less liquid bonds. Note however that our primary goal here is to demonstrate that a novel, alternative instance-based approach to explainability based on RF GAP proximities can help to uncover reasons for the model's success or failure, in a way that SHAP alone cannot.

\subsection{Data Description and Model Training}\label{sec:data_desc}
Here, we use an RF to generate estimations of the fair bid price of bonds across the US investment grade (IG) and high yield (HY) universes, every 15 minutes over the trading day. The performance of the model is then evaluated against executed trade prices, where the trade price is compared to the most recent 15-minute estimation for that security. 

The training label consists of executed trade returns, which are computed as the percent difference between the trade price and an estimate of the previous day's closing bid price for that security. The inputs to the model consist of 18 features, where all categorical variables in both train and test sets are label encoded based on label values in the training set. The main features include a coarse initial estimate of the security return, which is computed from a weighted average of trades and same-day quotes, together with 17 other features characterizing bond fundamentals and current market conditions. 
%1) a coarser initial estimate of the security return based in turn on an initial estimate of the security price, which is computed as a weighted average of trades and same-day quotes, 2) the average realized return for executed trades across the security's rating bucket over the past hour, 3) the average realized return for executed trades across the security's spread duration bucket over the past hour, 4) the average realized return for executed trades across the security's sector over the past hour, 5) ticker, 6) credit rating, 7) time to maturity, 8) duration times spread, 9) option-adjusted spread, 10) spread duration, 11) previous end of day price, 12) liquidity score (based on a quantile of the quotient of average daily volume and bid ask spread), 13) confidence score (a measure of the quantity and price dispersion of same-day quotes and trades), 14) 7-day rolling price volatility for the security, 15) whether the security is IG or HY, 16) industry classification, 17) hour of the day during which the estimation occurs, 18) sector classification.

The purpose of the random forest model here is to improve on the accuracy of this initial coarser price estimate by combining it with these other features. The trade data on which the model was trained was taken from the publicly available FINRA TRACE dataset, while the quotes were drawn from internal fixed income quote data. Bond characteristics such as sector, industry, ticker, duration, option-adjusted spread, rating, and maturity were also drawn from internal data sources.
%Description of how we are training random forest (train-test splits, cross-validations, imbalance data handling, which metrics we are using to evaluate random forest etc.

The model is trained on 9 months of public trade data for US investment grade (IG) and high yield (HY) corporate bonds, from Aug. 1 2022 to April 30 2023, and tested on the following three months. Hyperparameter tuning was performed using 5-fold walk-forward, expanding-window time series cross validation with randomized grid search over sklearn's RF regressor max\_depth, max\_features, min\_samples\_leaf, and n\_estimators parameters. To enable faster computation of the proximity matrix, the model was trained on a random sub-sample of 10,000 trades. 

Out of sample, the price estimate generated by the RF model showed a 9.07\% improvement in RMSE over the preliminary coarse estimate based on a weighted average of quotes and trades. For the RF, the test set RMSE was .356 as compared to .392 for the initial coarse estimate based on a weighted average of quotes and trades. 
%The train set RMSE for the RF was .246 as compared to .431 for the initial weighted average estimate.

\subsection{Results}\label{sec:results}

Here we consider several examples to show how GAP proximities can be used to explain the success or failure of an RF model on out-of-sample data points. 

\begin{figure}
\includegraphics[width=0.5\textwidth]{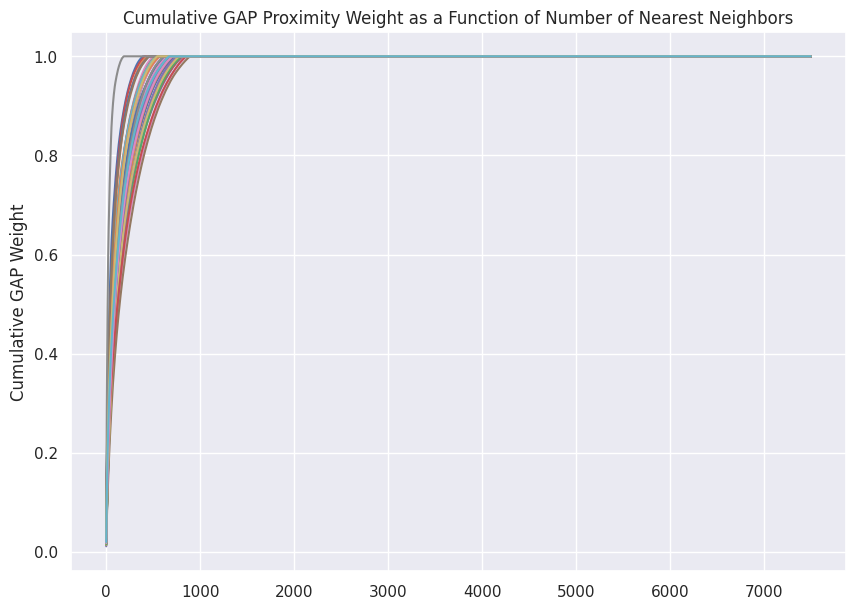}
\caption{Plotting cumulative GAP weight as a function of the number of nearest neighbors. On average, only 1/15 of the the total training set contributes to the RF model's prediction.}
\label{cumGapWeight}
\end{figure}

Taking a random sample of 50 data points from the test set, we can plot the cumulative GAP weight/proximity as a function of the number of nearest neighbors (sorted in descending order of weight/proximity) to check how many nearest neighbors are needed to generate a good approximation to the RF model's output. In Figure \ref{cumGapWeight}, we see that on average, only about 500 nearest neighbors out of the full training set of 7500 trades are needed to recover the original RF model prediction to close precision. 
%Of course, which specific points appear in this set will vary widely from one test point to another. 

\subsubsection{Explaining Model Performance Out of Sample}

\begin{figure}
\includegraphics[width=0.45\textwidth]{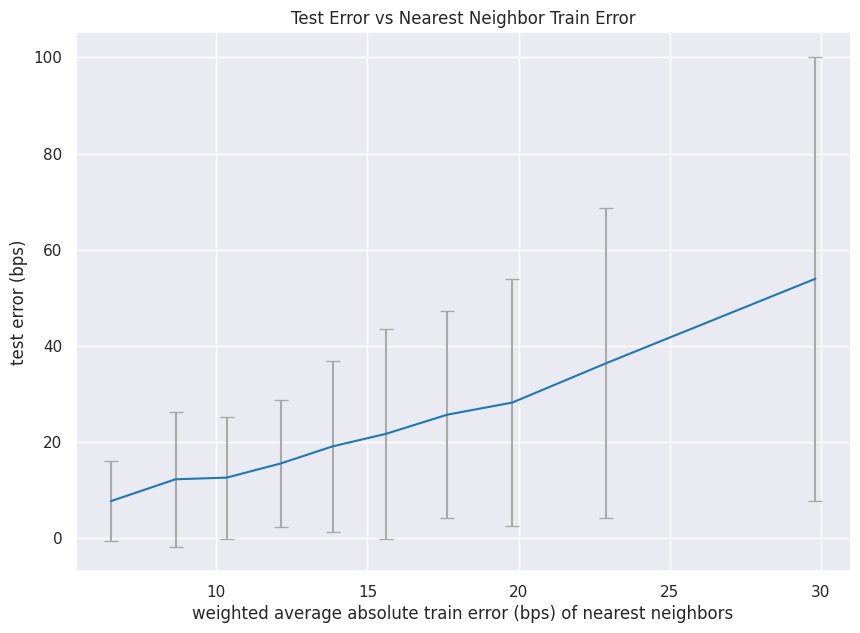}
\caption{Mean absolute test error as a function of the weighted MAE of the test point's nearest neighbors. The monotonic relationship indicates that higher proximity-weighted training error can help to explain why the model prediction was more likely to be off.}
%Correlation between test error and proximity weighted MAE, as measured per test point, is .49; between decile means of these errors it is .99. Thus, higher average training error for the nearest neighbors can help to explain why the model prediction was off for a given out-of-sample test case.}
\label{train_vs_test_error}
\end{figure}

One important application of proximities that we advance here is that they can be used to generate trust scores for the model predictions, which anticipate and explain the likely degree of error in a given RF prediction. For example, here we compute the proximity-weighted absolute error of training points, relative to the test point in question. Figure \ref{train_vs_test_error} shows the mean absolute error (MAE) out of sample as a function of this score. Since the MAE, as well as its standard deviation, tends to be larger when nearest-neighbor training errors are also large, we can often explain bad predictions in terms of the fact that the model's training error was also large in this region of the feature space. Beyond being an \textit{ex-post} explanation of model performance, the weighted mean absolute training error can serve as an \textit{ex-ante} measure of confidence in the model's predictions.

\begin{figure}
\includegraphics[width=0.4\textwidth]{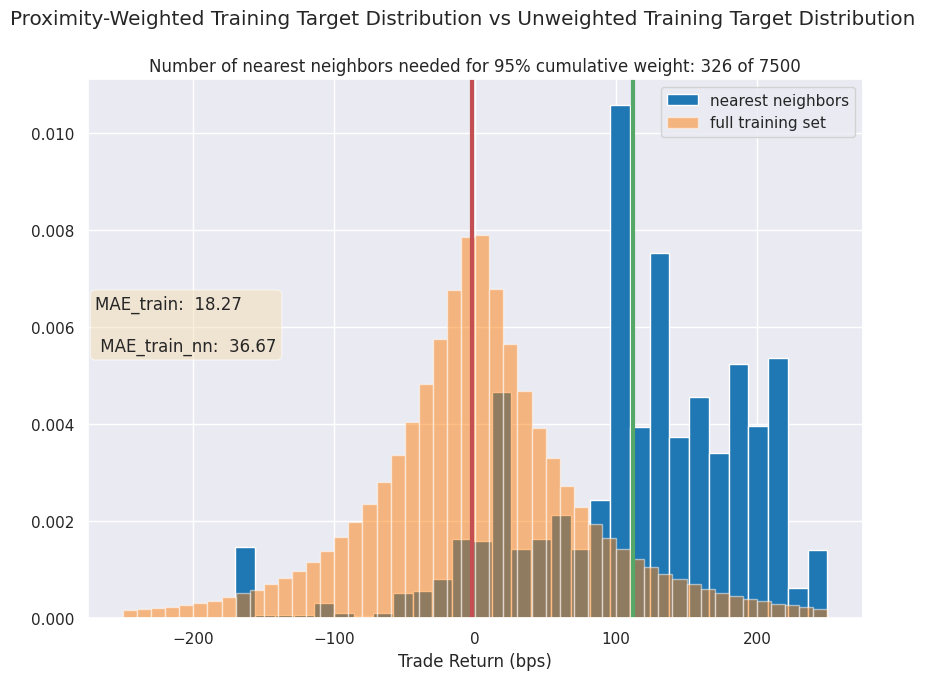}
\caption{Example, where out-of-sample model error is high, as predicted by the large value for the proximity-weighted train, set MAE.
%Two cases where the out-of-sample model error is high. 
%In both cases, the target labels (vertical red lines) have similar values and lie within the bulk of the full training distribution of target labels (orange histogram). However, in the upper plot, the subset of training points that the RF GAP proximity identifies as nearest neighbors (blue histogram) is clearly distinct from the subset of training points identified as nearest neighbors in the lower plot, helping to explain why the predictions (vertical green lines) are localized where they are in the two cases.
}
\label{badPred}
\end{figure}

\begin{figure}
\includegraphics[width=0.35\textwidth]{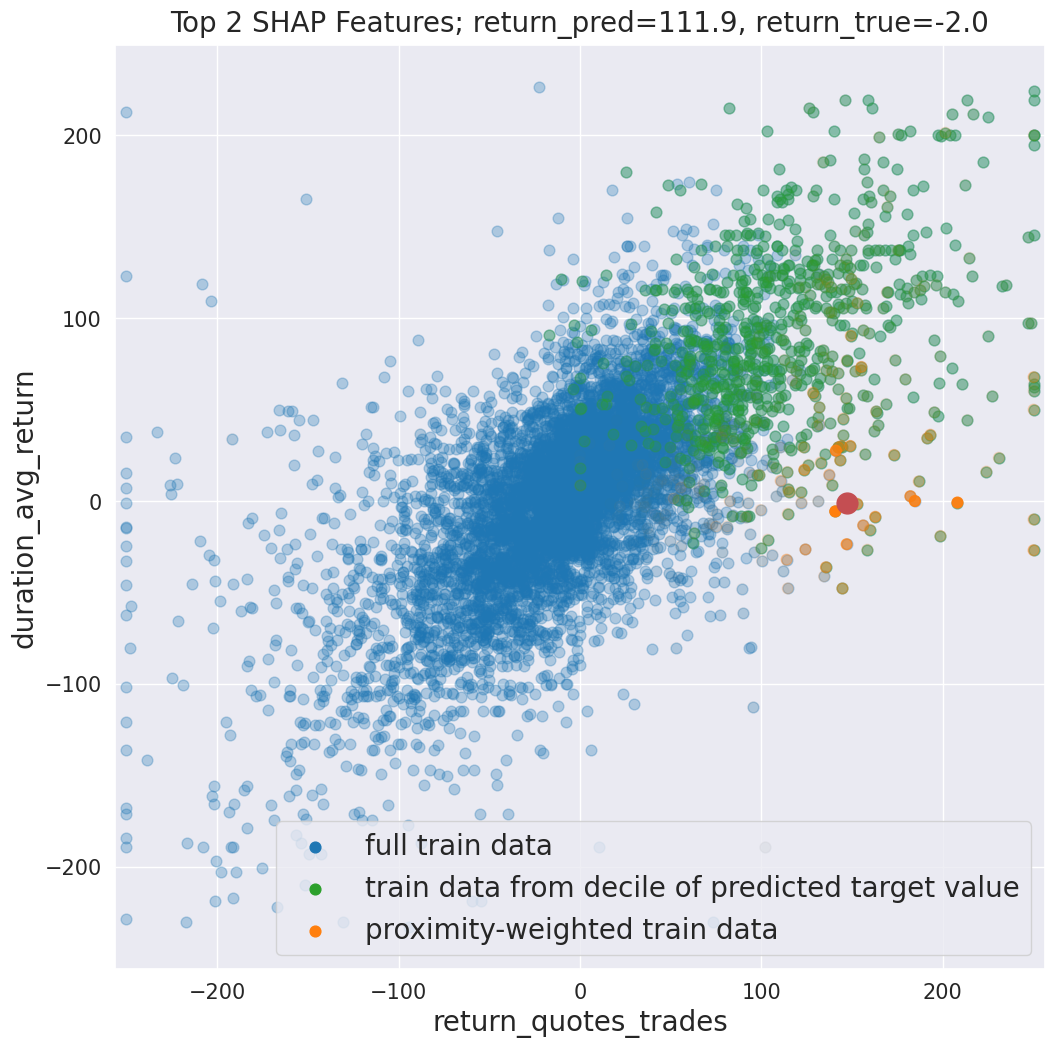}
\caption{Scatterplot of top 2 SHAP features at the test point (red). Green points are training points in the same decile bin of the predicted label as the test point. In orange are the training points with the largest proximity relative to the test point. The points most heavily influencing the model prediction are on the periphery of both the decile and the full training set, where data is sparse.  
%Plot of the two most important SHAP features for training data and test point (red dot). Training data is plotted according to the ground truth label, with blue indicating negative return and orange indicating positive. This plot is provided primarily as a basis for comparison with the lower plot. 
%\textbf{RIGHT:} The same scatterplot, but with opacity of points indicated in proportion to their proximity to the test point (red). High-weight training points contributing most to the prediction lie almost entirely on the periphery of the training distribution shown in the left diagram, indicating sparsity of training data in this region of the feature space. 
%This plot is the same as the one above, except that the proximity of each training point to the test point is indicated by its opacity. Here we see that the model prediction is most heavily influenced by the two orange points on either side of the test point.
}
\label{scatterplots_regression_badPred}
\end{figure}

\begin{figure}
\includegraphics[width=0.23\textwidth]{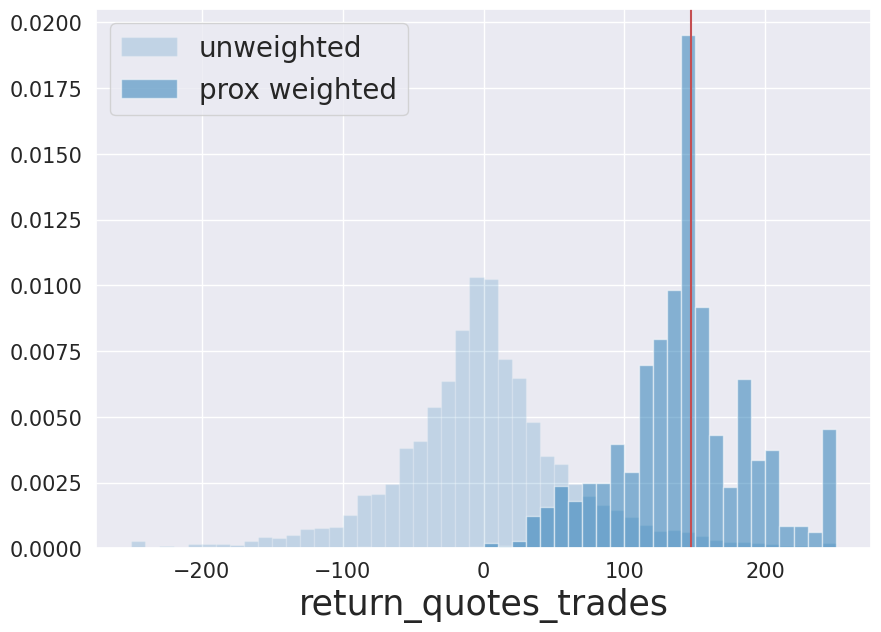}
\includegraphics[width=0.23\textwidth]{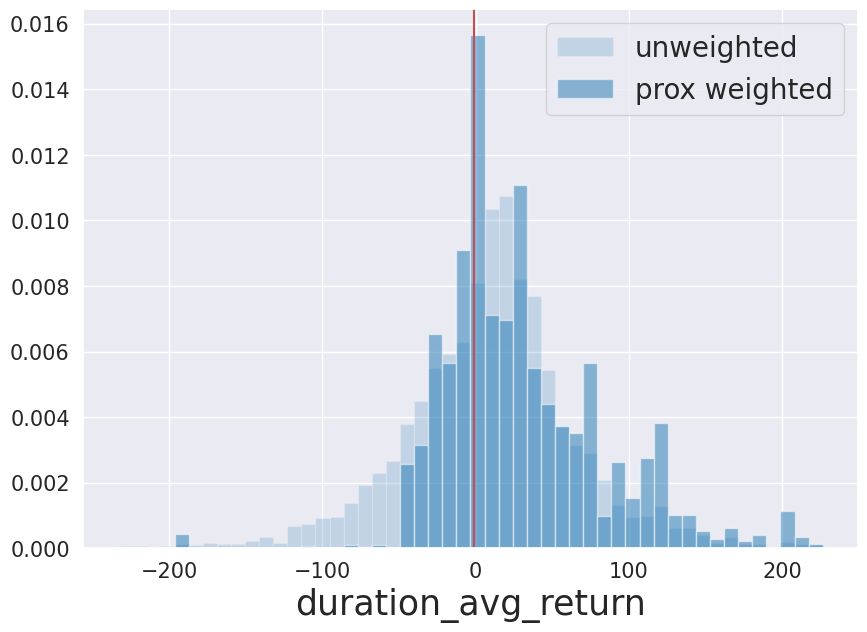}
\caption{Marginal plots showing separately the proximity-weighted vs unweighted train set feature distribution for each of the top 2 SHAP features (with the most important feature on the left). The proximity-weighted distribution of the most important feature (dark blue), is concentrated in the tails of the training distribution (faded blue).}
%As in the scatterplot above, we see that the value of the most important input feature for the test point lies in the periphery of the training distribution, where training data is relatively sparse, helping to explain why the prediction was inaccurate.}
%\caption{For each of the 3 most important SHAP features, starting from the top with the most important feature, marginal distributions of a) the proximity weighted training points from the predicted class, b) unweighted training points from both classes. Plots a) serve to show which regions of the feature space in this particular dimension contribute most to the model prediction. Plots b) show the overlap between the classes in this dimension.}
\label{marginal_plots_regression_badPred}
\end{figure}

\begin{figure}
\includegraphics[width=0.4\textwidth]{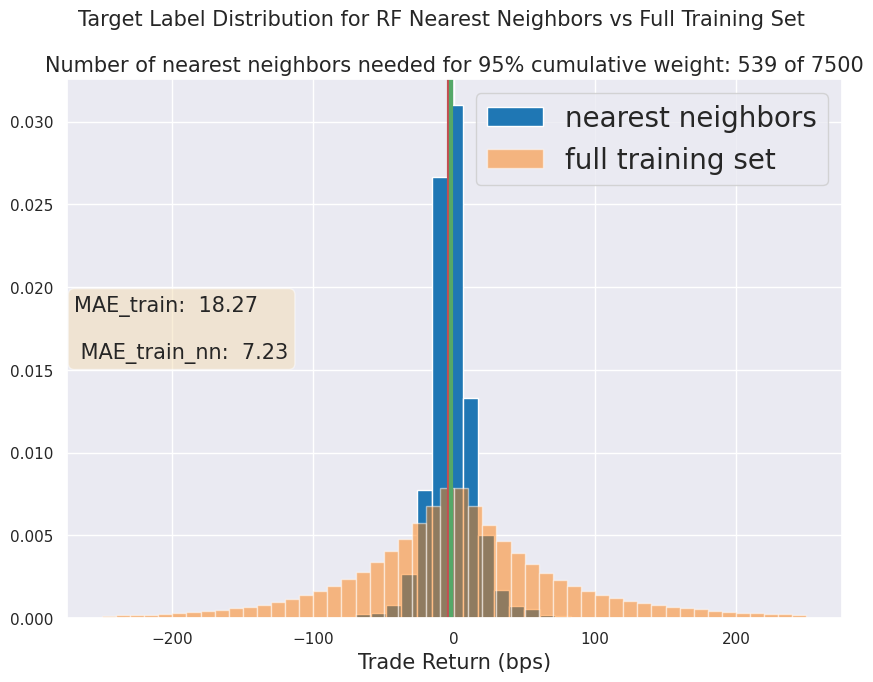}
\caption{Example where out-of-sample model error is low; note that the proximity-weighted absolute error on the train set is small compared with the training MAE.
%Two cases where the out-of-sample model error is low. 
%In both cases, the target label (vertical red line) lies close to the mean of the weighted distribution of train set nearest neighbor targets (vertical green line).
}
\label{goodPred}
\end{figure}

\begin{figure}
\includegraphics[width=0.5\textwidth]{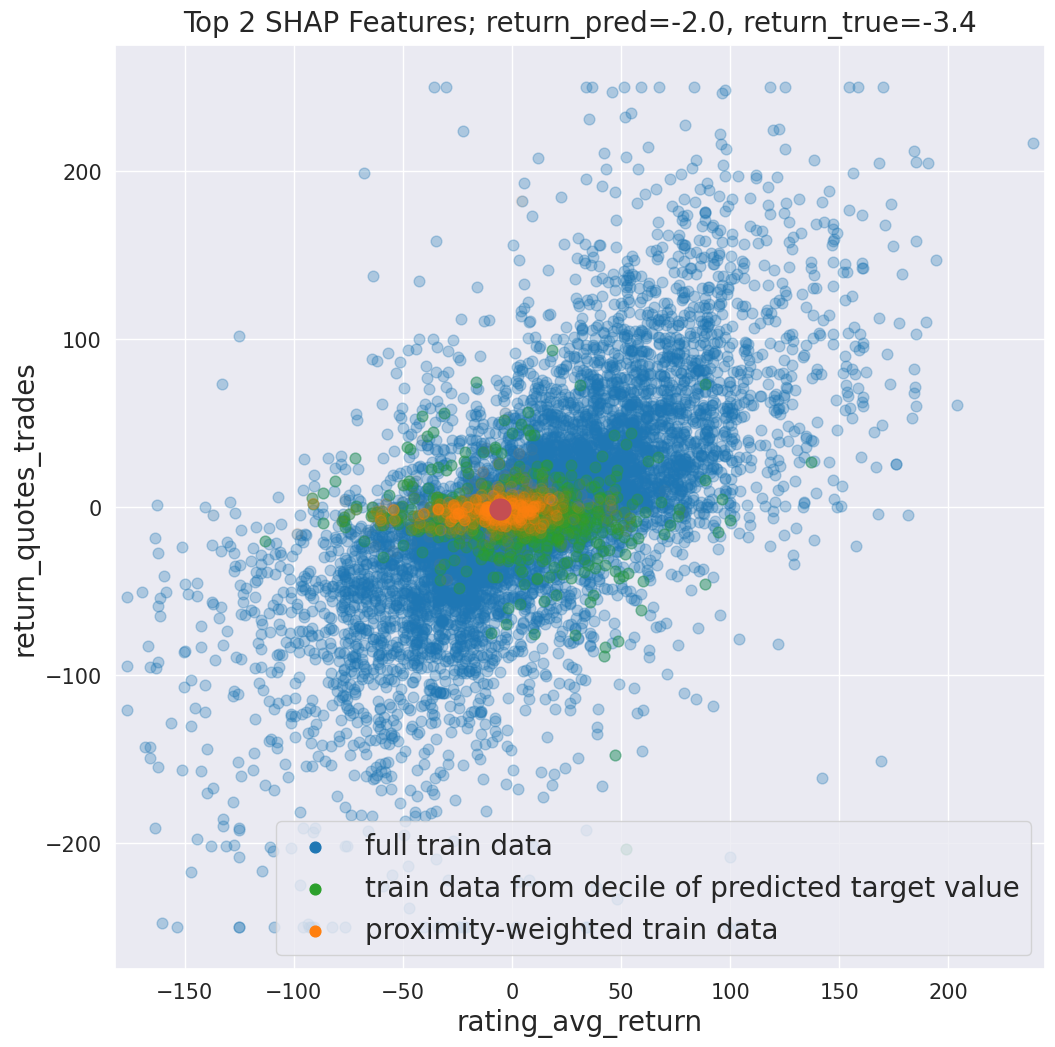}
\caption{Scatterplot of 2 most important SHAP features at the test point (red). In green are training points from the same decile of the model prediction. In orange are training points that contribute most to the model prediction, with opacity proportional to proximity to the test point.  
%Plot of the two most important SHAP features for training data and test point (red dot). Training data is plotted according to the ground truth label, with blue indicating negative return and orange indicating positive. This plot is provided primarily as a basis for comparison with the lower plot. 
%\textbf{RIGHT:} The same scatterplot of 2 most important SHAP features, but with opacity of points proportion to their proximity to the test point (red). Training points contributing most to the prediction lie well within the bulk of the training distribution shown in the left diagram, helping to explain why the prediction was more likely to be accurate. 
%This plot is the same as the one above, except that the proximity of each training point to the test point is indicated by its opacity. Here we see that the model prediction is most heavily influenced by the two orange points on either side of the test point.
}
\label{scatterplots_regression_goodPred}
\end{figure}

\begin{figure}
\includegraphics[width=0.23\textwidth]{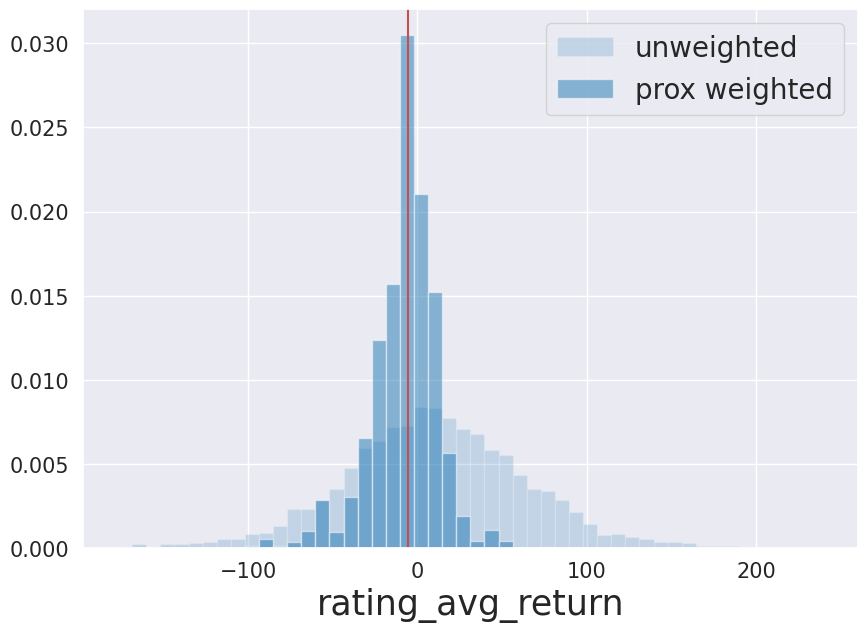}
\includegraphics[width=0.23\textwidth]{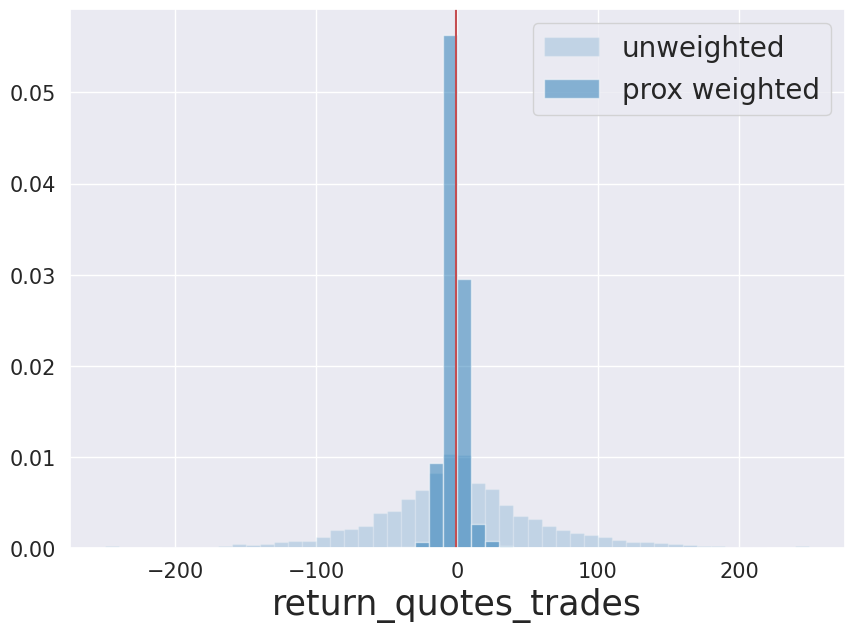}
\caption{Marginal plots showing proximity-weighted (dark) vs unweighted (faded) train set feature distribution for each of the top 2 SHAP features, with the most important on the left. With respect to the most important feature alone, the test point, along with the proximity-weighted distribution of this feature (dark blue), are concentrated in the bulk of the training distribution (faded blue) where data is plentiful.}
%\caption{For each of the 3 most important SHAP features, starting from the top with the most important feature, marginal distributions of a) the proximity weighted training points from the predicted class, b) unweighted training points from both classes. Plots a) serve to show which regions of the feature space in this particular dimension contribute most to the model prediction. Plots b) show the overlap between the classes in this dimension.}
\label{marginalplots_regression_goodPred}
\end{figure}

\subsubsection{Example Explanation for an Inaccurate Model Prediction}\label{sec:results}

Figure \ref{badPred} shows an out-of-sample observation where the model's prediction of the return was far from the realized return in the trade. The proximity-weighted distribution of target variables at the test point (blue histogram) furnishes not only an estimate of the conditional mean $E(y|x)$ of the target variable given the inputs, but of the entire conditional predictive distribution $p(y|x)$, where $E(y|x) = \int dx y p(y|x)$. Compared to the case of a successful prediction in \ref{goodPred}, it is clear visually that the predictive distribution here is much more spread out than is typical for the more successful prediction, advising a lower degree of confidence in the model's prediction in this case. Similarly, comparing the proximity-weighted MAE of nearest neighbors to the train set MAE shows that in this case, the model is likely to be substantially less accurate than is typical. As shown in Figure \ref{badPred}, the GAP-proximity-weighted mean absolute training error of the test observation's nearest neighbors, 36.67, was unusually high as compared with the mean absolute error for the entire training set, 18.27. 

%The histograms show that even though in both cases the realized trade return (vertical red line) was well within the distribution of realized trade returns seen in the training set (orange histogram), this return falls well outside the distribution of trade returns for the nearest neighbor training points (blue histogram) that were responsible for generating the RF prediction (vertical green line), where a 95\% cumulative weight threshold was used to determine the number of nearest neighbors to include. Moreover, we can see why the model was more likely to be off in these cases by virtue of the training errors for the test point's nearest neighbors. As shown in Figure \ref{badPred}, the GAP-proximity-weighted mean absolute training error of the test observation's nearest neighbors, 36.67, was unusually high as compared with the mean absolute error for the entire training set, 18.27. 

Turning to the input features rather than the target, Figs \ref{scatterplots_regression_badPred} and \ref{marginal_plots_regression_badPred} show that the test point (red), and the proximity-weighted training points that contributed most to the model prediction at this point (orange), lie in a region of feature space where data is relatively sparse, helping to explain why the prediction was off.

\subsubsection{Example Explanation for an Accurate Model Prediction}

Figure \ref{goodPred}, by contrast, shows an example where the model's prediction of the return was close to the realized return in the trade. The figure shows that the realized return (vertical red line) lies close to the mean (vertical green line) of the proximity-weighted distribution of realized returns (blue histogram). As in the case of the inaccurate predictions, we can see in this case why the model was more likely be accurate, by virtue both of the substantially more narrow predictive distribution and of the nearest neighbor training errors. As compared to an MAE of 18.27 for the entire training set, the proximity-weighted MAE for the test point was a much lower 7.23. 

Turning to the input features, Figs \ref{scatterplots_regression_goodPred} and \ref{marginalplots_regression_goodPred} indicate that the test point, and the proximity-weighted training points that contributed most to the model prediction at this point, lie squarely within the bulk of the training distribution for the most important features, explaining why the model was more likely to be accurate in this case.

\section{Classification Use Case: Predicting the Direction of Realized Trade Returns}

To illustrate the proximity-based approach to model explanation in the context of classification with random forests, we adapt the above model to predict only the \textit{direction} of the return of the next trade relative to the previous end-of-day price. This is a binary classification problem, although the methods described here are straightforwardly extended to problems with three or more classes. Out of sample, the binary classifier that we trained had f1 score=.81, accuracy=.81, precision=.80, and recall=.81 relative to the positive class (corresponding to positive return).  

\subsection{Results}
\begin{figure}
\includegraphics[width=0.5\textwidth]{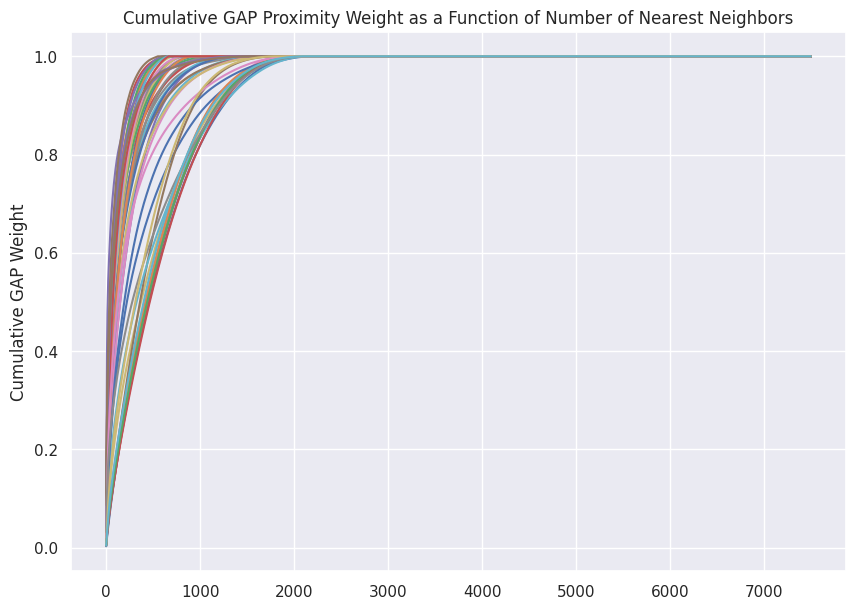}
\caption{\textbf{Cumulative GAP weight as a function of number of nearest neighbors, for binary classification.} }
\label{cum_gap_weight_clf}
\end{figure}

Here, as in the regression case, we seek to explain both the prediction and the out-of-sample performance of the model based on its nearest neighbors in the training set, where nearest neighbors are those training points with the largest GAP proximities - i.e., with the largest weights $k_{ij}(x)$ in the expansion (\ref{eq:RF_weighted_KNN_clf}). Figure \ref{cum_gap_weight_clf} shows for $50$ random test points the number of nearest training neighbors that are needed to recover the prediction of the random forest to a given precision. Note that in the case of binary classification, the number of nearest neighbors contributing to the RF prediction on a given test point is larger, ostensibly because the target label takes on only 2 values rather than a continuum.  

As in the regression case, we illustrate with two example explanation, one of an inaccurate prediction, and one of an accurate prediction.

\subsubsection{Explaining Model Performance Out of Sample}

Two ways in which one might explain the poor performance of a KNN classifier on a given test point are 1) the values of the target variable for nearest neighbors in the training set show high variability around this region in the feature space, 2) the test point is relatively isolated in the sense that on average it has low proximity to points in the train set, indicating that training data is sparse in the vicinity of the test point. The proximity-based class probability can be low even in the bulk of the data distribution, while the outlier score ratio can only be low in regions of the feature space where data is relatively sparse.

\begin{figure}
\includegraphics[width=0.45\textwidth]{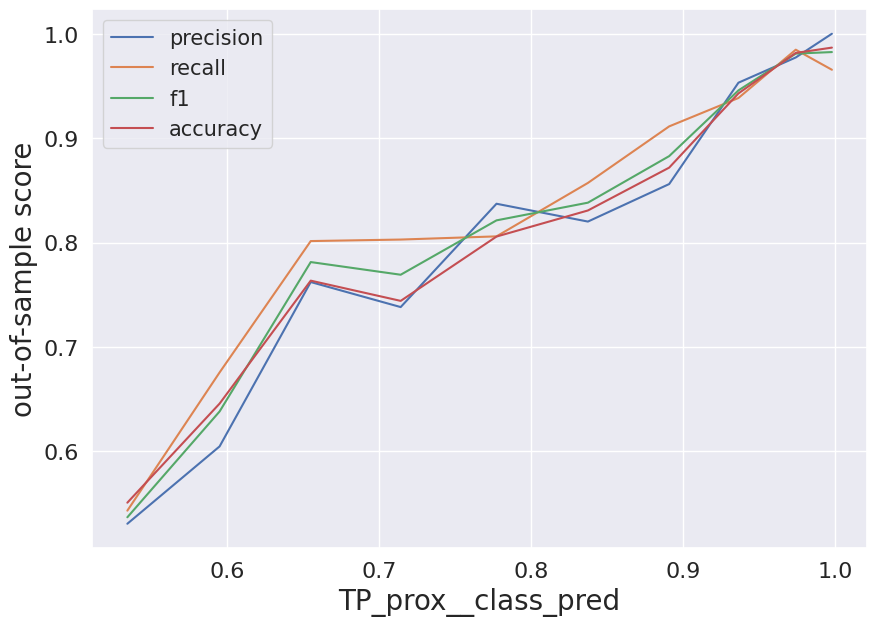}
\caption{Out-of-sample performance as a function of the proximity-based probability for the predicted class.}
%All four measures of out-of-sample performance increase essentially monotonically with this score, which helps both to give a \textit{ex post} explanation of out-of-sample performance as well as an \textit{ex ante} trust score anticipating the model's likely performance out of sample.}
\label{score_vs_TPprox}
\end{figure}

Concerning the first type of explanation, Figure \ref{score_vs_TPprox} shows the dependence of the out-of-sample-classification performance as a function of the proximity-based class probability. Here we see a clear monotonic relationship between the out of sample performance and the proximity-weighted class probability. On this basis, we can understand why a given out-of-sample prediction was more or less likely to be correct based on the proximity-weighted performance of the model on the test point's nearest neighbors in the training set, and thereby furnish an explanation for the out-of-sample performance of the model in a given instance. 

Concerning the second type of explanation, random-forest proximities can also be used to define an outlier measure for a given test point with respect to each class, based on how ``far'', in terms of proximities, the members of that class tend to be from the test point. This indicates how typical a given point is of points from the same class in the training set. Here, we follow \cite{desai2023quantifying} in using the GAP proximities to define an outlier measure with respect to each class. 

\begin{figure}
\includegraphics[width=0.45\textwidth]{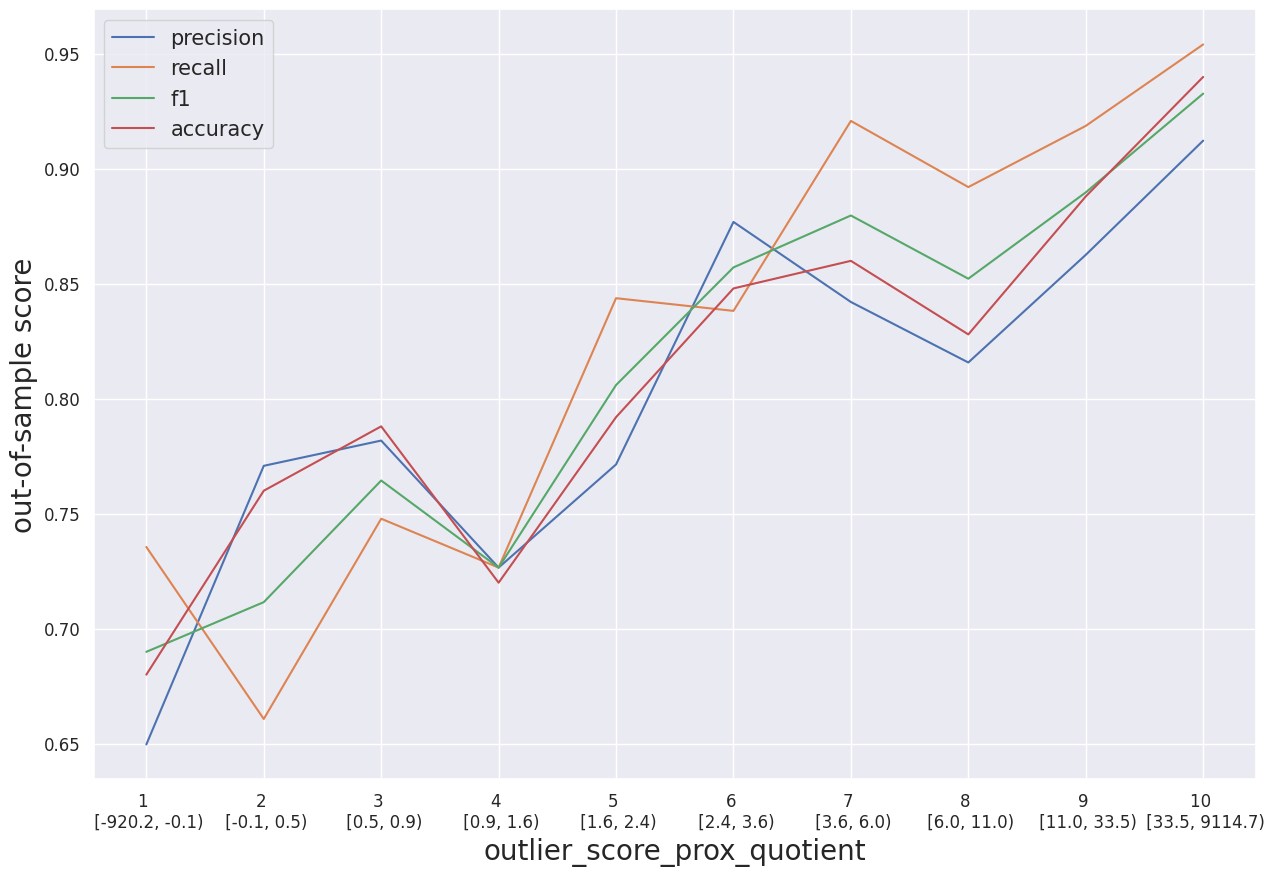}

\caption{
%\textbf{TOP:} Out-of-sample performance (precision, recall, f1-score, accuracy) as a function of the proximity-based outlier score of the test point, with respect to the class that was \textit{not} predicted; note that performance is indicated with respect to deciles in this quantity, with associated bin intervals indicated along the bottom tier of labels along the x-axis.
%Intuitively, this score reflects the sparsity of training data for the class that was not predicted. We see  that as this score gets large, we can rule out the non-predicted class with greater certainty and gain greater confidence that the test point will turn out to belong to the predicted class. However, we also see that for smaller values of this quantity, the out of sample performance rises slightly, as one would expect since this scenario reflects a case where the test point lies in the center of the training data distribution, and therefore should have improved performance. 
%\textbf{BOTTOM:} 
Out-of-sample performance with respect to the ratio of proximity-based-outlier scores between the non-predicted and predicted class; decile bin intervals are indicated along the bottom tier of labels along the x-axis. 
%Intuitively, this score reflects the \textit{relative} sparsity of training data around the test point between the class that was not predicted and the class that was predicted. Here we see that out-of-sample performance improves more monotonically (with some noise) as the score increases.
}
\label{score_vs_outlier_quotient}
\end{figure}

This outlier measure is defined as follows. First, for a given test point, the raw outlier score of the test point with respect to class $J$ is defined in terms of the average proximity of the test point to all points from class $J$ in the training set. This average proximity is given by
\begin{equation}\label{RF_outlier_score_clf_1}
P^{J}(i) \equiv \sum_{j \in cl(J)} Prox^{2}(i,j),
\end{equation}
\noindent where $Prox(i,j)=Prox^{GAP}(i,j)$, and the sum is over all points with target label $J$ in the training set; henceforth we will drop the $GAP$ superscript in denoting the GAP proximities. To facilitate comparison across classes with different numbers of training points, we further normalize by the number of training points $n_J$ in class $J$, giving the average proximity per training point in class  $J$. The raw outlier score is then defined as the inverse of the average proximity per training point:
\begin{equation}\label{RF_outlier_score_clf_2}
O^{J}_{raw}(i) \equiv \frac{n_j}{P^{J}(i)}.
\end{equation}
To compute the final outlier score $O^{J}(i)$ for the test point $i$ with respect to class $J$, we compute a median-based z-score of the raw outlier score, using the median raw outlier score $med_{J}(O^{J}_{raw}(i))$ with respect to points in class $J$ as the measure of central tendency (in place of the usual mean $\mu$), and the median absolute deviation from the median, $dev_{J}(O^{J}_{raw}(i))$ as the measure of spread (in the place of the usual standard deviation $\sigma$):
\begin{equation}\label{RF_outlier_score_clf_3}
O^{J}(i) \equiv \frac{O^{J}_{raw}(i) - med_{J}(O^{J}_{raw}(i))}{dev_{J}(O^{J}_{raw}(i))}.
\end{equation}
This outlier score provides a more principled measure of typicality with respect to its class of a test point's input features than do measures based on Euclidean distance, based as it is on a notion of distance/similarity that is defined specifically to take account of each feature's relevance to predicting the class label. Figure \ref{score_vs_outlier_quotient} shows the dependence of the out-of-sample classification performance as a function of the proximity-based outlier score. 
\footnote{More precisely, it shows the dependence of the out-of-sample performance on the deciled ratio between the outlier score with respect to the not-predicted class and the outlier score with respect to the predicted class. 

\footnote{In the case of three or more classes, one can take this other class to be the class that is next-closest in terms of median proximity to the test point.} On this basis, we can understand why a given out-of-sample prediction was more or less likely to be correct based on its relative position in feature space with respect to training points from both classes and on the relative sparsity of training data for the two classes.}

Fig \ref{score_vs_outlier_quotient} shows the out-of-sample performance as a function of the ratio between the outlier score for the class that was not predicted and the one that was predicted, indicating again a clear monotonic dependence, which shows that the score is indeed a good predictor of performance. 

%Figure \ref{score_vs_outlier_nonpred} shows the out-of-sample performance as a function of training data sparsity not in a relative sense, as measured in terms of the ratio of the outlier scores for the two classes, but in an absolute sense, as measured in terms of the outlier score directly. Here we see that for the most part, out-of-sample performance improves as the outlier score of the class that was \textit{not} predicted increases. However, we see that the reverse is true at the low end of the outlier score range, as a small outlier score for the non-predicted class can reflect that the test point lies in a region of the feature space where there is a lot of training data from both classes, and thus where the performance is likely to be better. 

We can obtain a more detailed understanding of the model performance for each test point by exploiting the decomposition of the two trust scores above in terms of the nearest neighbors to the test point, where ``near" is understood in terms of the GAP proximities. For the proximity-based class probability score, this decomposition is given by (\ref{eq:RF_weighted_KNN_clf}), while for the proximity-based outlier score, it is given by (\ref{RF_outlier_score_clf_1}), (\ref{RF_outlier_score_clf_2}), and (\ref{RF_outlier_score_clf_3}).

\subsubsection{Example Explanation for an Inaccurate Model Prediction}

%Above, we see that random forest proximities can be used to define two distinct measures of trustworthiness for the model, one based on local variability in the target variable, and another based on the sparsity of training data for each class in the vicinity of the test point. 

\begin{figure}
\includegraphics[width=0.23\textwidth]{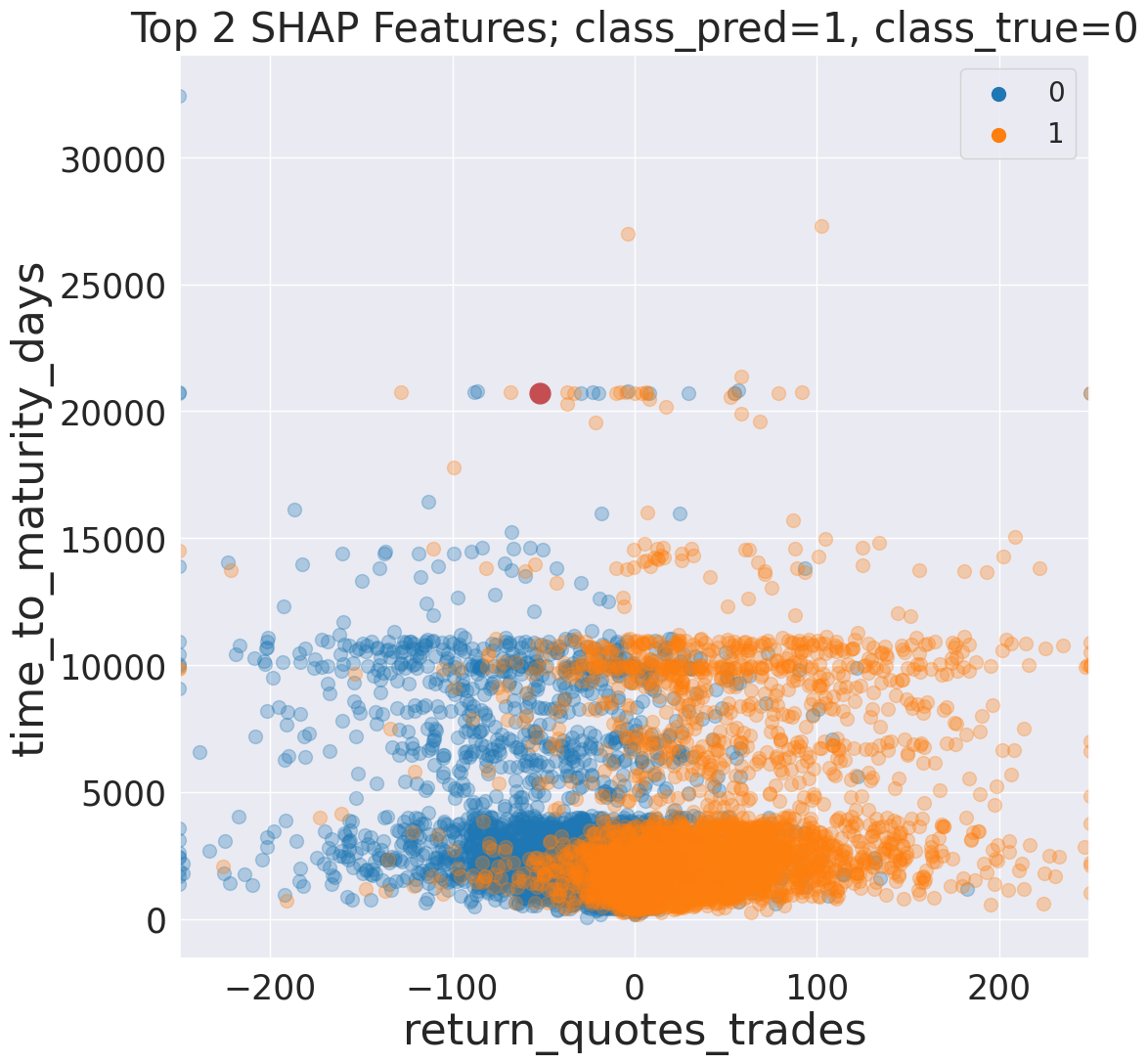}
\includegraphics[width=0.23\textwidth]{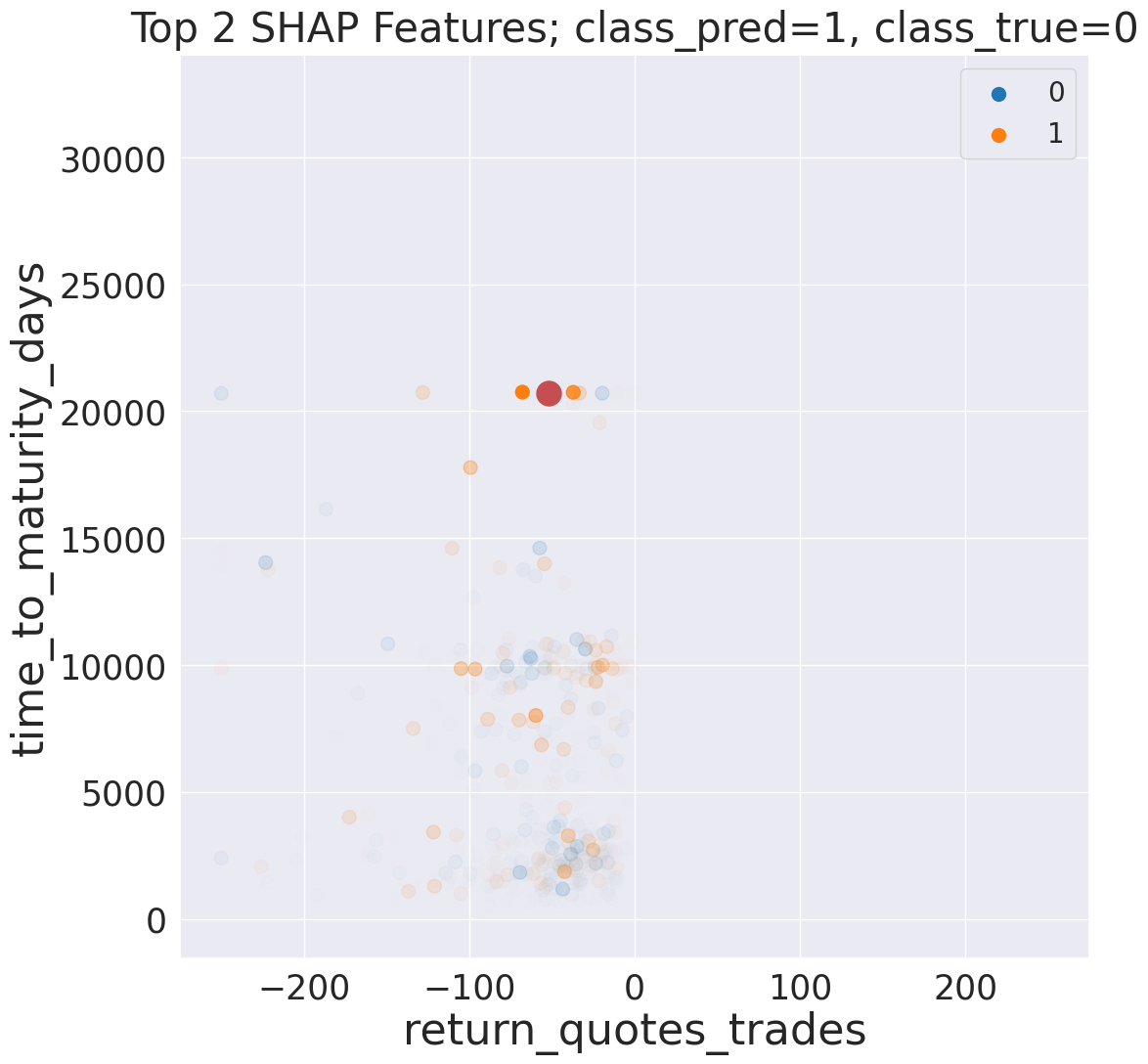}
\caption{\textbf{LEFT:} Training distribution for the two most important SHAP features, with blue/orange indicating negative/positive return, and the red dot designating the test point. \textbf{RIGHT:} Same, except that the proximity of each training point to the test point is now indicated by its opacity. Here the model prediction is primarily influenced by the two orange points on either side of the test point.} 
\label{scatterplots}
\end{figure}

\begin{figure}
\includegraphics[width=0.23\textwidth]{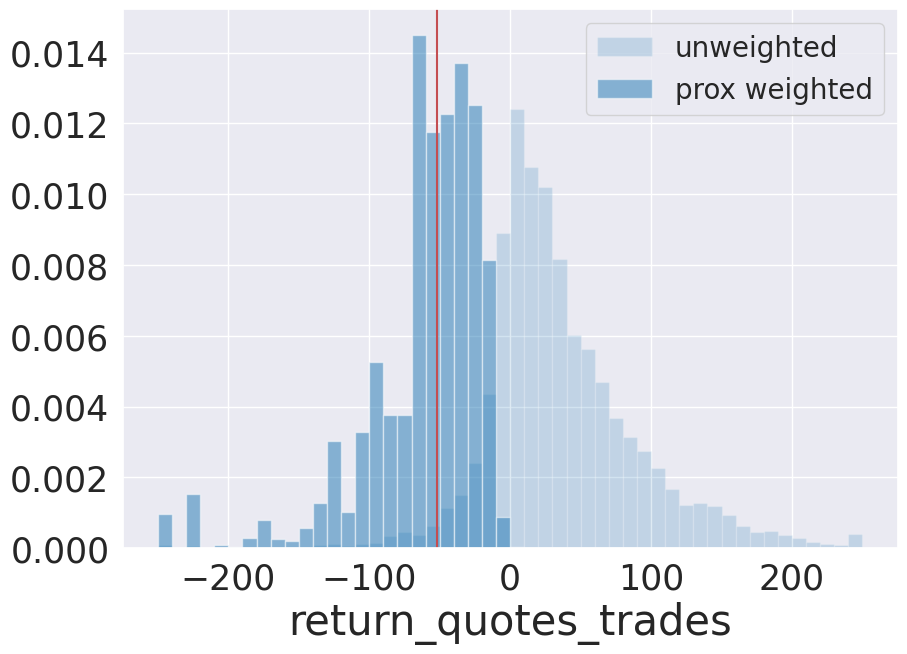}
\includegraphics[width=0.23\textwidth]{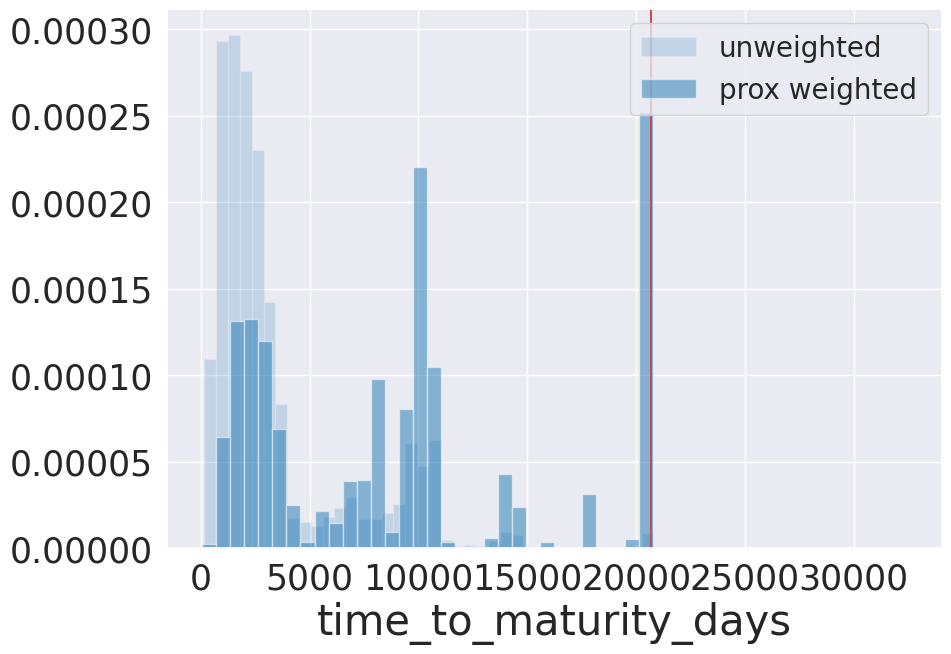}
\caption{Marginal plots for each of the top 2 SHAP features (with most important on the left), showing proximity-weighted vs unweighted train set feature distribution for the predicted class.}
%, b) unweighted train set feature distribution for both classes).} 
%\caption{For each of the 3 most important SHAP features, starting from the top with the most important feature, marginal distributions of a) the proximity weighted training points from the predicted class, b) unweighted training points from both classes. Plots a) serve to show which regions of the feature space in this particular dimension contribute most to the model prediction. Plots b) show the overlap between the classes in this dimension.}
\label{marginalplots}
\end{figure}

Considering first an example where the model prediction was off, Fig \ref{scatterplots} plots the test point (red) and training points (orange for positive return, blue for negative) in terms of the two most important SHAP features. In the left plot of each figure, training points from the two classes are shown without indication of their weight in the RF prediction. In the lower plot of each figure, the weight is now indicated by the opacity of each training point. In Figs \ref{marginalplots} and \ref{marginalplots_correct} are shown two separate plots of the marginal distribution for each of the top two SHAP features, with the left plot corresponding to the most important feature; each plot contrasts the proximity-weighted distribution of values of the feature (dark blue) with the unweighted distribution for this feature (light blue), for training points with ground truth label in the predicted class. %The lower plot in each pair contrasts the unweighted distributions over values of this feature for the two classes. 

Fig \ref{scatterplots} shows that the proportion of training points that contribute to this prediction is relatively small. In fact, in the right plot, we see that the two orange points on either side of the test point contribute far more strongly to the model outputs than any other points in the training set, explaining why the prediction generated by the model was for a positive return. We can also see in the marginal plots of Fig \ref{marginalplots} why the model was relatively likely to be off in this case: for the second most important feature (and actually also the third, which is not shown), the test point lay on the periphery of training distribution, indicating sparsity of training data associated with the predicted class in the vicinity of the test point. Likewise, the most important feature shown on the left, also shows that the local proximity-weighted distribution is dramatically altered from the unconditional training distribution for this feature, heavily over-weighting a small number of training points on the periphery of the dataset.  

 Comparing the values for this test point of the proximity-weighted probability of the predicted class, $.62$, and of the quotient of the proximity-based outlier scores, $-5.02$, to the plots of these scores against the out-of-sample performance in Fig \ref{score_vs_TPprox} and \ref{score_vs_outlier_quotient}, respectively, we see in both cases that the scores do in fact anticipate relatively poor performance. 

\subsubsection{Example Explanation for an Accurate Model Prediction}

\begin{figure}
\includegraphics[width=0.25\textwidth]{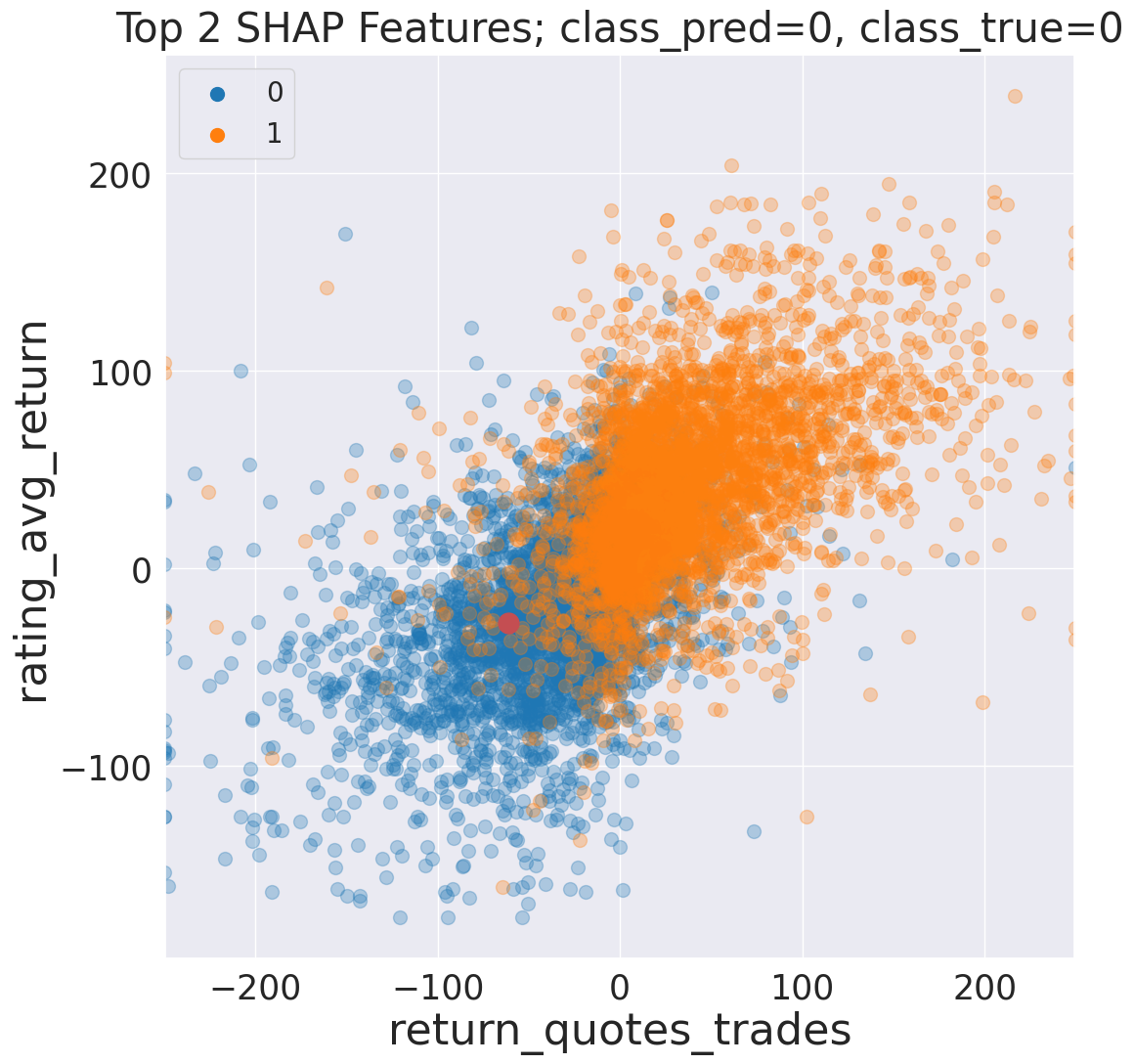}\includegraphics[width=0.25\textwidth]{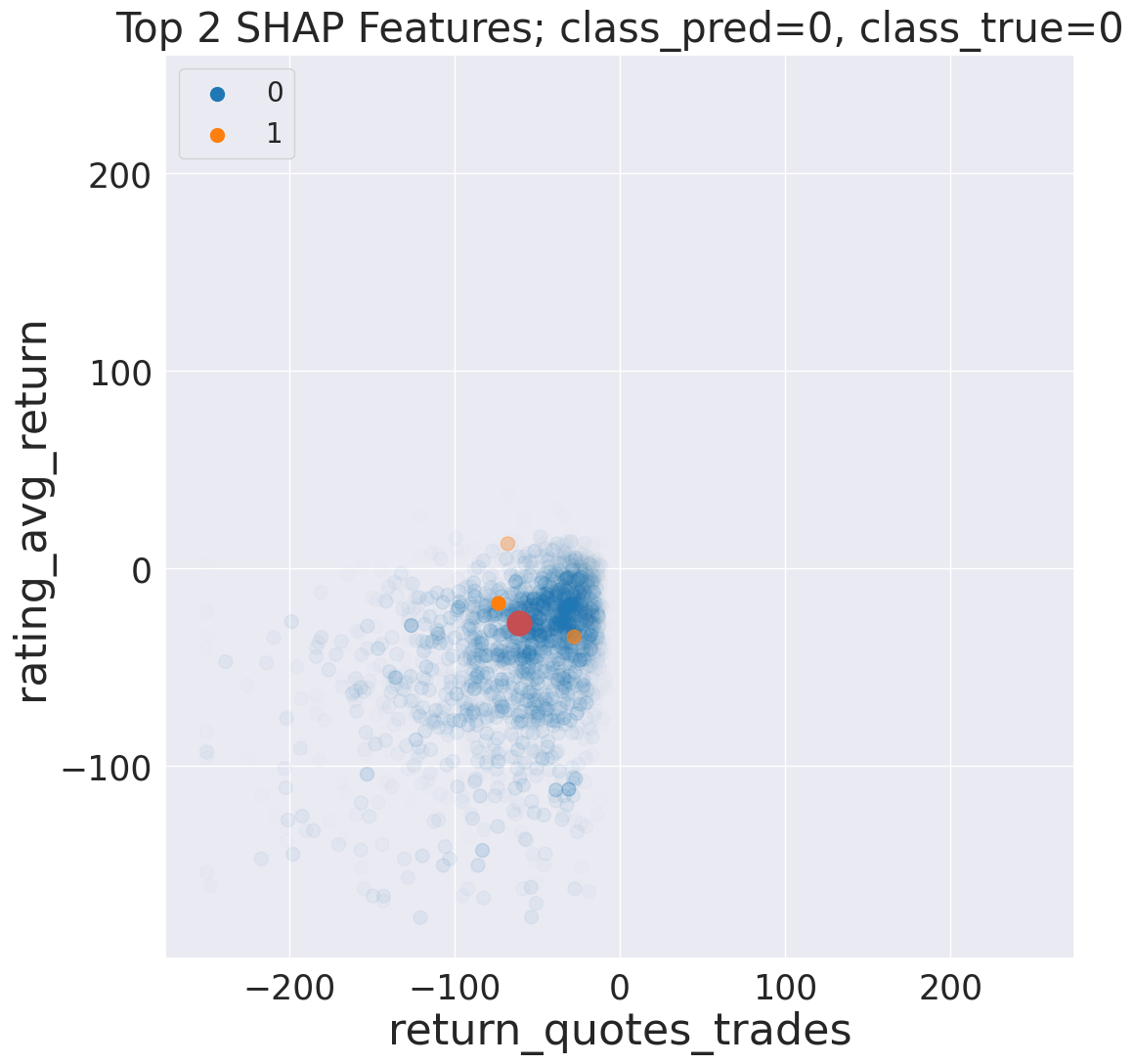}
\caption{\textbf{LEFT:}  
Training distribution for top 2 SHAP features, with blue/orange indicating negative/positive return, and red dot designating the test point.  \textbf{RIGHT:} Same, except the proximity of each training point to the test point is indicated by its opacity. The model prediction is generated by training points belonging mostly to the predicted class (blue).}
%Scatterplot of the two most important SHAP features for training data and test point (red dot). Training data is plotted according to the ground truth label, with blue indicating negative return and orange indicating positive. \textbf{RIGHT:} Same as scatterplot above, except that the proximity of each training point to the test point is indicated by its opacity. Here we see that the model prediction was most heavily influenced by a relatively large neighborhood of training points, almost all of which belong to the predicted class.} 
\label{scatterplots_correct}
\end{figure}

\begin{figure}
\includegraphics[width=0.23\textwidth]{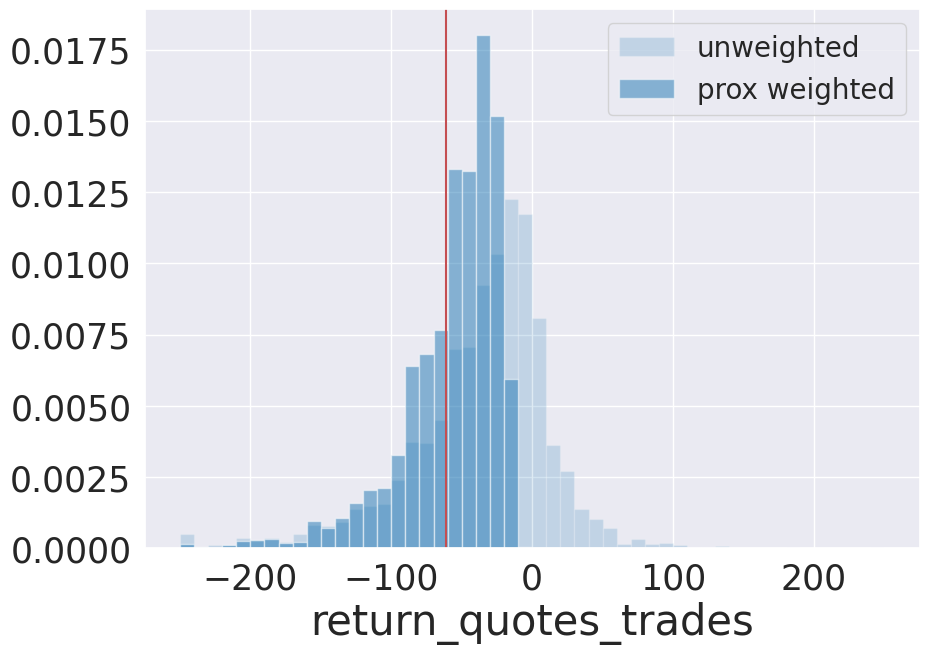}
\includegraphics[width=0.23\textwidth]{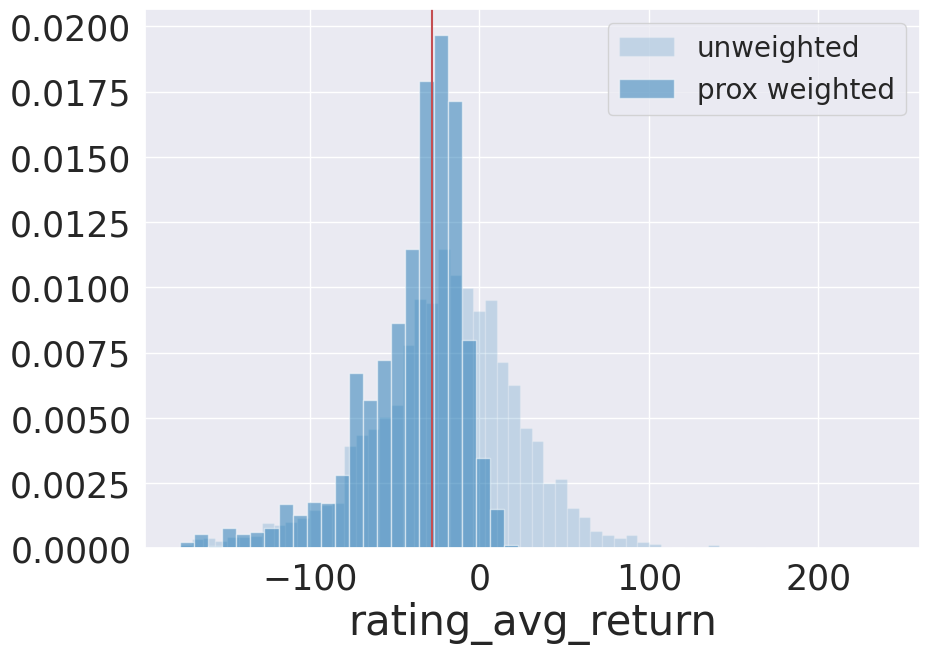}
\caption{Proximity-weighted vs unweighted marginal feature distribution for points from predicted class, for each of the top 2 SHAP features (with most important to the left). The test point and proximity-weighted distribution both fall within the bulk of the training distribution. }
%b) unweighted train set feature distribution for both classes).} 
\label{marginalplots_correct}
\end{figure}

Turning to an example of a successful prediction, Figs \ref{scatterplots_correct} and \ref{marginalplots_correct} depicts a case in which the model correctly predicted negative trade return. By comparison with the previous example, we see in Fig \ref{scatterplots_correct} that the proportion of training points contributing significantly to the prediction is relatively large. In addition, most of these points are associated with a negative realized return, explaining why the model predicted negative return. The marginal plots for the most important feature, shown in the two plots of Fig \ref{marginalplots_correct}, confirm that the test point lies in a region of the feature space where almost all of the training points have negative returns.

 Comparing the values for this test point of the proximity-weighted probability of the predicted class, $.98$, and of the quotient of the proximity-based quotient of outlier scores, $5.18$, to the plots in Fig \ref{score_vs_TPprox} and \ref{score_vs_outlier_quotient}, respectively, we see in both cases that out-of-sample performance is anticipated to be relatively good.

\section{Conclusion}\label{sec:conclusion}

The instance-based approach to random forest model explanation based on GAP proximities complements and enhances SHAP explanations in several important ways. First, it allows for explanation not just of the model predictions, but of the model performance out of sample. Proximities identify cases where the model is more likely to be off, either because the target variable shows greater variability and uncertainty in the region around the test point, or because the data is relatively sparse. This type of explanation provides not just an \textit{ex-post} but also an \textit{ex-ante} measure of trust in the model predictions. Although we have illustrated these claims in the context of corporate bond pricing, the use of GAP proximities for the explanation of out-of-sample model performance can be extended to generic applications of random forests. A similar application of proximities is also available in the context of GBMs.

%While SHAP has the advantage of being model agnostic, instance-based approaches to model explanation have the advantage that they provide a strategy not only for attributing model predictions, but for explaining why these predictions were more or less likely to be close to the realized target label. Moreover, this latter type of explanation provides an ex-ante measure of confidence in the model predictions. By incorporating information about the training error of nearest neighbors, instance-based explanations based on the GAP proximity take account of facts about the varying level of noise in the target variable across different regions of the feature space, and thus can also help explain why the out of sample error of the model prediction was relatively high or low.

\section{Acknowledgement}
The views expressed here are those of the authors alone and not of BlackRock, Inc.
% BlackRock, Inc.
%BlackRock, Inc.
%\bibliographystyle{unsrt}

\bibliographystyle{ACM-Reference-Format}
\bibliography{sample-base}
\end{document}